\documentclass{article}

 \usepackage[dblblindworkshop, final]{neurips_2025}
 \usepackage{graphicx}

\usepackage[utf8]{inputenc} 
\usepackage[T1]{fontenc}    
\usepackage{hyperref}       
\usepackage{url}            
\usepackage{booktabs}       
\usepackage{amsfonts}       
\usepackage{nicefrac}       
\usepackage{microtype}      
\usepackage{xcolor}         

\title{SoDaDE: Solvent Data-Driven Embeddings with Small Transformer Models}
\workshoptitle{Machine Learning and the Physical Sciences}

\author{%
  Gabriel Kitso Gibberd \\
  Imperial College London \\
  London, UK
  \And 
  Jose Pablo Folch \\
  SOLVE Chemistry \\
  London, UK
  \And
  Antonio Del Rio Chanona \\
  Imperial College London \\
  London, UK
}

\begin{document}

\maketitle

\begin{abstract}
Computational representations have become crucial in unlocking the recent growth of machine learning algorithms for chemistry. Initially hand-designed, machine learning has shown that meaningful representations can be \textit{learnt} from data. Chemical datasets are limited and so the representations learnt from data are generic, being trained on broad datasets which contain shallow information on many different molecule types. For example, generic fingerprints lack physical context specific to solvents. However, the use of harmful solvents is a leading climate-related issue in the chemical industry, and there is a surge of interest in green solvent replacement. To empower this research, we propose a new solvent representation scheme by developing Solvent Data Driven Embeddings (SoDaDE). SoDaDE uses a small transformer model and solvent property dataset to create a fingerprint for solvents. To showcase their effectiveness, we use SoDaDE to predict yields on a recently published dataset, outperforming previous representations. We demonstrate through this paper that data-driven fingerprints can be made with small datasets and set-up a workflow that can be explored for other applications.
\end{abstract}

\section{Introduction}


Chemistry needs better representations for molecular modelling \citep{yang2019analyzing}. Widely-used representations \citep{landrum2006rdkit, probst2022reaction} capture general molecular features rather than structures and properties relevant to the modelling task. Increasingly complex data-driven representations \citep{ahmad2022chemberta, schwaller2021mapping} are helping address these issues, however, they are not guaranteed to generalise, especially to unseen or niche chemistry. 

As a clear example, we address the issue of solvent representation highlighted by the newly proposed Catechol Benchmark \citep{boyne2025catechol}. Most chemical reactions both in industry and academia occur in solvents. Many solvents are volatile organic compounds (VOCs) which contribute to environmental issues, such as smog formation, and health problems including cancer \citep{jindamanee2025environmental}. The exploration for greener and safer solvents is ongoing but work is needed to model solvent impact accurately and identify replacements for harmful VOC solvents.

In this work we create a new data-driven fingerprint for solvents. Dubbed Solvent Data-Driven Embeddings (SoDaDE), we identify a small but suitable solvent property dataset and train a transformer model \citep{vaswani2017attention} via data augmentation. We assess performance on the pre-training task and then showcase the usefulness of the new fingerprint by outperforming previous representations on the Catechol Benchmark \citep{boyne2025catechol}. We provide data and code \footnote{\url{https://github.com/g-a-b-r-e-a-l/SoDaDE_Solvent_Data-Driven_Embeddings_from_Language_Models}}.

\section{Background and Related Work}

\subsection{Common Chemical Representations}
For human understanding, molecular representations are strings of characters, like colloquial names, IUPAC names and Simplified Molecular Input Line Entry System (SMILES) strings \citep{weininger1988smiles}. For machine understanding, molecular simulations and representations are used. Molecular simulations, such as density functional theory (DFT) calculate molecular behaviour, but complexity scales exponentially with size so simulations struggle with large molecules \citep{goedecker1999linear}. 

Molecular representations allow machine learning from chemical datasets by communicating a molecule to a machine. Representations have two main categories, molecular graphs \citep{duvenaud2015convolutional} and rule-based fingerprints, although descriptors, like DFT values are also used \citep{ahneman2018predicting}. A molecular graph is a network, with nodes as atoms and edges as bonds. These show close connectivity well but struggle with global structure and usability \citep{szymkuc2016computer}. Rule-based fingerprints are more accessible and often used for chemical property (ECFPs) \citep{rogers2010extended} and yield prediction (DRFPs) \citep{probst2022reaction}. These fingerprints one-hot encode each substructure present into a long vector. Consequently, rule-fingerprints communicate structure well but are unspecific and not information dense. 

\subsection{Related Work}
Based on the success of language foundation models \citep{devlin2019bert, touvron2023llama}, similar ideas have emerged to attempt the building of chemical foundation models. From this, data-driven molecular fingerprints (DDfps) have risen to address the issues with common chemical representations \citep{honda2019smiles}. DDfps are representations learnt by pre-training a neural network, commonly a transformer model, on related data through self-supervised learning. Then the second last layer, before an output is calculated, is used as a representation. If trained correctly, these vectors will contain information \citep{pratt1992discriminability} about the molecules and lead to better results when fine-tuned.

ChemBERTa \citep{chithrananda2020chemberta, ahmad2022chemberta}, based on BERT \citep{devlin2019bert}, is an early molecule DDfp trained on 77 million SMILES strings and showed good performance in property prediction. Reaction Fingerprints (RXNFP) \citep{schwaller2021mapping, schwaller2021prediction} pre-trained on reaction classification and achieved excellent performance on yield prediction. Recently, T5-Chem \citep{lu2022unified} explored pre-training on multiple tasks simultaneously. However, due to the range of molecular types in large datasets, DDfps tend to perform worse on niche prediction tasks.

\subsection{Semantic Understanding of Chemical Data}
Different transformer architectures \citep{vaswani2017attention}) are used for fingerprint generation and most learn from text, like  ChemBERTa and RXNFP, which rely on tokenising SMILES strings. However, chemical datasets often include numerical data so data-types need to be combined to leverage the entire dataset. We pre-train with a similar scheme to \citet{yin2020tabert}, who project text and tabular data into the same embedding space.

\section{Data and Method}

\citet{boyne2025catechol} recently proposed a new machine learning benchmark, named the Catechol Benchmark, for solvent selection. They measured the yields of the rearrangement of allyl substituted catechol under different reaction conditions, combining a variety of solvents, temperatures and reaction times. From their relatively poor results, they concluded that the machine learning community needs better solvent representations.

Out of the representations they explored, the solvent descriptors found in \citet{spange2021reappraisal} performed the best. This aligns with chemical understanding; solvent properties are known to affect reaction success, so we focus our transformer model's pre-training on this dataset. A few example rows can be found in Table \ref{tab:Spange-dataset-transposed}.. The solvent-property relationships gained from pre-training should create a better solvent featurisation to improve upon the Catechol Benchmark results.

\begin{table*}[h!]
\centering
\small
\begin{tabular}{llllllllllll}
\toprule
Solvent & Type & ET(30) & $\alpha$ & $\beta$ & $\pi^*$ & SA & SB & SP & SdP & $n$ & $\delta$ \\
\midrule
n-pentane & alkane & 31 & 0 & 0 & -0.08 & 0 & 0.073 & 0.593 & 0 & 1.358 & 14.5 \\
3-methylpentane & alkane & -- & -- & -- & -- & 0 & 0.05 & 0.62 & 0 & -- & -- \\
\bottomrule
\end{tabular}
\caption{Example data from the Spange solvent dataset. It contains 191 solvents, their type, and molecular properties. Some values are missing, so we leverage the ability of transformers to mask these values during training. The full set of properties are ET(30): Reichardt's polarity parameter, $\alpha$: hydrogen donating ability, $\beta$: hydrogen accepting ability, $\pi^*$: Kalmel-Taft polarisability parameter, SA: solvent acidity, SB: solvent basicity, SP: solvent polarisability,  SdP: solvent dipolarity, $N_{mol\ cm^{-3}}$: molecular density, $n$: solvent refractive index, $f(n)$: a function of $n$ quantifying the non-specific solute-solvent interactions, $\delta$: a correction term for polarisability.}
\label{tab:Spange-dataset-transposed}
\end{table*}

\begin{figure*}[htb]
  \centering
  \includegraphics[width=\textwidth]{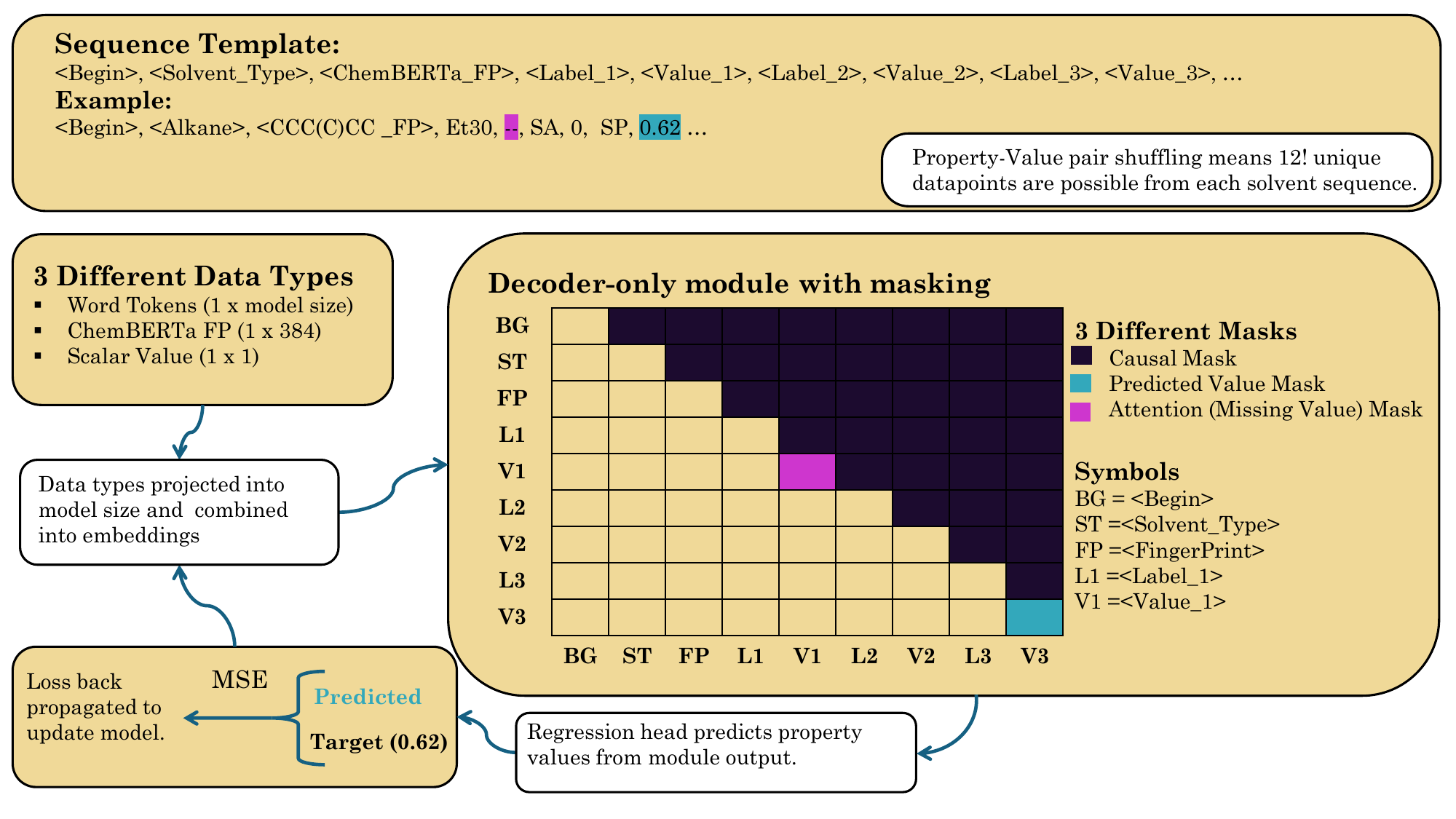}
 \caption{This diagram shows the data augmentation and model training structure. The Spange descriptors (top of the diagram) used to construct solvent sequences which could be shuffled to create more data points. The model combines the items in these sequences into the standardised model size vectors. The decoder then learns from these sequences, predicting any values hidden by the mask token, based on the previous sequence (causal mask) and ignores missing values (attention mask).}
  \label{fig: data_and_model_structure}
\end{figure*}

The Spange solvent property set contains up to 12 molecular properties of 191 solvents. While a small dataset, it is appropriate as commonly used solvents are limited. We augment the dataset size by creating `solvent sequences', which is achieved by randomly shuffling property-value pairs in the sequence a maximum number of 12! combinations per solvent. For training, mask tokens were used to cover-up random solvent properties. We learn a model which attends to preceding, property values and predicts the properties covered by the mask token. A causal mask, common in transformer models, was used to ensure the model focussed on tokens before the token to be predicted. Finally, an attention mask was used to hide missing values from the model. A summary of the data processing and training method is illustrated in Figure \ref{fig: data_and_model_structure}.

Choices within the pre-training task have a significant impact on fine-tuning performance \citep{liu2023same} so the validation and test set were chosen based on the solvent types in the Catechol dataset. For each of the 9 solvent types in the Catechol dataset, 1 random, complete solvent was taken from the Spange dataset and added to the validation set. This incentivised the transformer model to learn good representations of the relevant solvent types. For the test set, we took 5 additional solvents from the Spange set, one for each of the most common solvent types in the Catechol set to get a proxy of performance on the relevant downstream task.

Model parameters were explored using grid search with the help of a high-performance cluster. Final model parameters for the SoDaDE model were a dimension size of 64, 16 attention heads, 5 layers and a hidden dimension of 4 $\times$ 64. A learning rate scheduler was used which started at a learning rate of 0.001 and was reduced by a factor of 0.5 every 5 epochs of no improvement until it reached a learning rate of 1e-8. It used a mask token rate of 0.3 and achieved a final, normalised MSE of 0.107 on the validation set. For fine-tuning, on the Catechol benchmark task, the SoDaDE decoder output of the last token was used as the solvent representation. This was passed through one small multi-layer perceptron (MLP) and then combined with the temperature and residence time in a second MLP to predict the percentage starting material, product 2 and product 3. We test two versions of our model, one where we fix the SoDaDE weights (i.e. without fine-tuning), and another when we allow for changing of the weights when training on the data (i.e. with fine-tuning).

\section{Results}
\subsection{Solvent Property Prediction}

While not a guarantee of success, performance on a relevant pre-training task is a good indicator that the model is learning effective representations \citep{liu2023same}. In Table \ref{tab:predictions_metrics} performance of SoDaDE is compared with a Gaussian process (GP) model, Random Forest (RF) model, and the average of each property in the training set (AVG). For the competing models, we featurize solvents using molecular fingerprints calculated using RDKit \citep{landrum2006rdkit, rogers2010extended}.

\begin{table*}[h!]
\centering
\caption{This table compares the performance of different methods on predicting properties of the 5 solvents in the test set using MSE.}
\begin{tabular}{@{}lcccc@{}}
\noalign{\hrule height 0.8pt}
Metric & Averaged Values & RF Predictions & GP Predictions & SoDaDE Predictions \\
\noalign{\hrule height 0.4pt}
ET30 & 40.06 & 4.00 & 1.33 & \textbf{1.22} \\
$\alpha$ & 0.115 & \textbf{0.0004} & 0.0050 & 0.0010 \\
$\beta$ & 0.0750 & 0.0237 & 0.0143 & \textbf{0.0084} \\
$\pi^{\star}$ & 0.112 & 0.0515 & \textbf{0.0057} & 0.0170 \\
SA & 0.0201 & 0.0001 & 0.0012 & \textbf{0.00009} \\
SB & 0.0617 & 0.0280 & 0.0144 & \textbf{0.0020} \\
SP & 0.0083 & 0.0033 & \textbf{0.0008} & 0.0012 \\
SdP & 0.0848 & 0.0294 & \textbf{0.0128} & 0.0160 \\
$N_{mol\ cm^{-3}}$ & 4.00e-6 & 2.00e-6 & \textbf{1.00e-6} & \textbf{1.00e-6} \\
n & 0.0050 & 0.0013 & 0.0006 & \textbf{0.0003} \\
fn & 0.0013 & 0.0003 & 0.0002 & \textbf{0.00007} \\
$\delta$ & 10.2 & 2.46 & 2.64 & \textbf{0.771} \\
\noalign{\hrule height 0.8pt}
Average MSE & 4.23 & 0.550 & 0.335 & \textbf{0.170}\\
\noalign{\hrule height 0.8pt}
\end{tabular}
\label{tab:predictions_metrics}
\end{table*}

\subsection{Performance on the Catechol Benchmark}

The Catechol Benchmark is split into two different tasks, one of predicting reaction outcomes with a single solvent, and a second "full dataset" task. The full data involves predicting reaction outcomes under multiple solvent mixtures. To account for this, weighted average of the two solvent fingerprints were taken and fed to the neural network. The model was adapted to a fork of the Catechol GitHub to ensure consistent testing. These results are displayed in Table \ref{tab:Catechol Benchmark results}. 

We compare against the ACS solvent selection guide's PCA representation \citep{diorazio2016toward}, the differential reaction fingerprints \citep{probst2022reaction}, RDKit's molecular fragments and fingerprints concatenated (fragprints) \citep{landrum2006rdkit, griffiths2022data}, \citet{spange2021reappraisal}'s featurization with GP imputation, reaction fingerprints \citep{schwaller2021mapping, schwaller2021prediction}, and ChemBERTa fingerprints \citep{ahmad2022chemberta}. These are all the methods compared against in the original benchmarking paper \citep{boyne2025catechol}. Interestingly, we achieve stronger performance in the full data as opposed to the single solvent data. This suggests we have learnt a good representation for solvent mixtures, which original methods struggled to do well. 

\begin{table}[h!]
\centering
\caption{Comparison of the performance of different methods on the Catechol benchmark. The competing method performances are taken from the original paper. Their repository was forked and SoDaDE was added to ensure consistency in the task. The values are the MSE between the predicted yield values and actual values of the starting material, and two products as percentages.}
\begin{tabular}{@{}lccc@{}}
\noalign{\hrule height 0.8pt}
Model & Featurisation & MSE Full Data ($\downarrow$) & MSE Single Solvent ($\downarrow$) \\
\noalign{\hrule height 0.4pt}
MLP & ACS & 0.0140 & 0.0110 \\
& DRFPs & 0.0130 & 0.0150 \\
& Fragprints & 0.0110 & 0.0100\\
& Spange (GP imputation) & 0.0100 & 0.0100\\
\noalign{\hrule height 0.4pt}
LLM & RXNFP & 0.1050 & 0.0550\\
& ChemBERTa & 0.1530 & 0.0740\\
& SoDaDE (ours) without tuning & 0.0029 & \textbf{0.0044}\\
& \textbf{\textit{SoDaDE (ours) with tuning}}& \textbf{0.0026} & \textbf{0.0044} \\
\noalign{\hrule height 0.8pt}
\end{tabular}
\label{tab:Catechol Benchmark results}
\end{table}

\section{Conclusion}

The SoDaDE model shows significant improvements over tested methods, demonstrating that we have created an effective solvent fingerprint. The similarity between the non-finetuned and fine-tuned models shows this fingerprint is not dependant on fine-tuning and captures general solvent features. Through SoDaDE, we demonstrate that effective fingerprints can be created from small datasets and invite others to replicate our method within their own domains.

\bibliographystyle{unsrtnat}
\bibliography{references}

\newpage


\appendix

\section{Investigation of SoDaDE parameters}

Deep neural networks are models that lack significant interpretability. They capture complex relationships in a way that is difficult to extract, visualise, or  understand. This lack of interpretability is a loss for chemistry, where the relationships found could contribute significantly to chemical understanding.

\begin{figure*}[htbp]
  \centering
  \includegraphics[width=\textwidth]{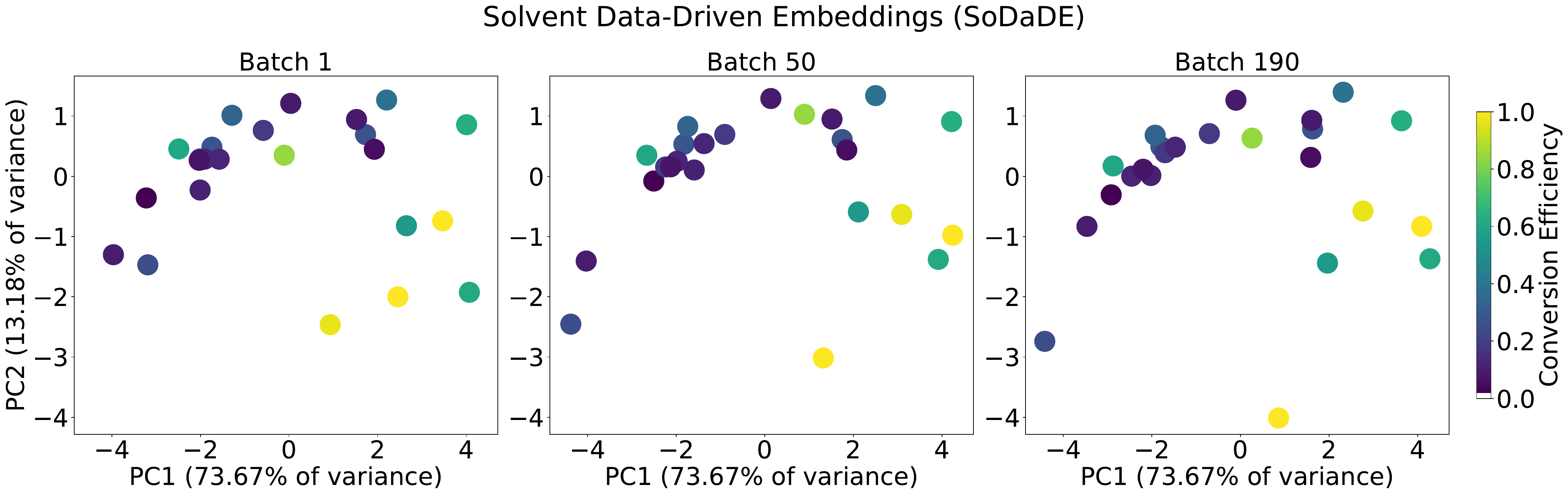} \\[1em]
  \includegraphics[width=\textwidth]{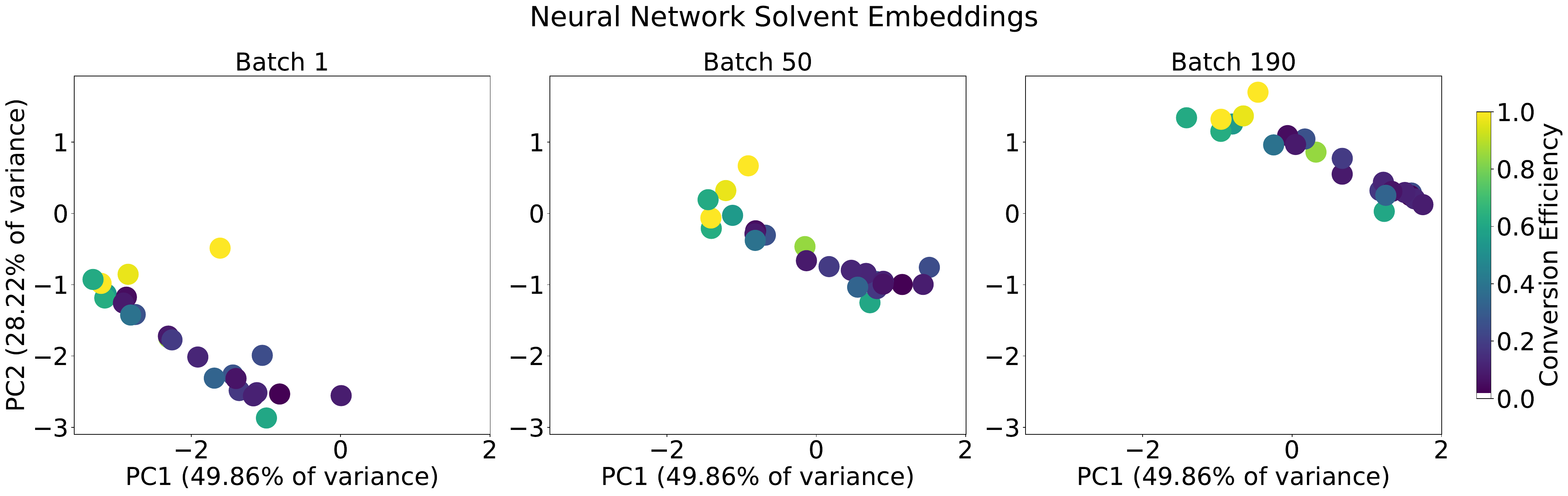}
  \caption{This shows the transformation of the solvent embeddings during training, plotted in PCA-reduced space from 64 dimensions down to 2. The neural network (NN) embeddings organise as training continues, moving from the lower left corner to the upper right. The SoDaDE embeddings are disordered and do not organise noticeably. The low learning rate of the SoDaDE model likely contributes to this, but the performance achieved suggests the model creates effective solvent fingerprints.}
  \label{fig: Solvent Embeddings}
\end{figure*}

Figure \ref{fig: Solvent Embeddings} plots the solvent embeddings from the SoDaDE and NN models over 200 batches of training. These embeddings have been reduced from 64-bit vectors to 2 bits using principal component analysis (PCA). The colour map values use the conversion efficiency, Equation \ref{eq: conversion efficieny},  of the catechol reaction in each solvent. To calculate conversion efficiency, the maximum quantities of Products 2 and 3 and the minimum quantity of SM were used. Colour mapping this way achieves better distinction between solvents.

\begin{equation}
\label{eq: conversion efficieny}
    Value = \frac{(Product 2 + Product 3)}{(Product 2 + Product 3 + SM)}
\end{equation}

\end{document}